\documentclass{article}

\usepackage{neurips_2026}
\usepackage[utf8]{inputenc} 
\usepackage[T1]{fontenc}    
\usepackage{hyperref}       
\usepackage{url}            
\usepackage{booktabs}       
\usepackage{amsfonts}       
\usepackage{nicefrac}       
\usepackage{microtype}      
\usepackage{natbib}
\usepackage{graphicx}
\usepackage{tcolorbox}
\tcbuselibrary{skins,breakable}
\usepackage{amsmath}
\usepackage{amssymb}
\usepackage{mathtools}
\usepackage{algorithm}
\usepackage{algorithmic}
\usepackage{wrapfig}
\usepackage{amsthm}
\theoremstyle{plain}
\newtheorem{theorem}{Theorem}[section]

\theoremstyle{definition}

\theoremstyle{remark}

\usepackage{graphicx}
\usepackage{booktabs}
\usepackage{multirow}
\usepackage[textsize=tiny]{todonotes}
\usepackage[utf8]{inputenc}
\usepackage{caption}
\usepackage{enumitem}
\usepackage{nicematrix}
\usepackage{booktabs}
\usepackage[table,xcdraw]{xcolor}
\usepackage{arydshln}

\newtcolorbox{promptbox}[1]{
    enhanced,
    colback=gray!5!white,       
    colframe=black!75!white,    
    coltitle=white,             
    fonttitle=\rmfamily\bfseries, 
    fontupper=\rmfamily\small,          
    toptitle=5pt,                
    bottomtitle=5pt,             
    left=12pt, right=12pt, top=5pt, bottom=5pt, 
    arc=1.5mm,                  
    boxrule=0.8pt,
    title=#1
}

\usepackage{times} 

\title{
  Whispers in the Noise: Surrogate-Guided Concept Awakening via a Multi-Agent Framework
}

\author{%
  Mengyu Sun$^{1,2}$\thanks{These authors contributed equally.} \quad
  Ziyuan Yang$^{3}$\footnotemark[1] \quad
  Zunlong Zhou$^{2}$ \quad
  Junxu Liu$^1$ \quad
  Haibo Hu$^1$ \quad
  Yi Zhang$^2$ \\
  \\
  $^1$Department of Electrical and Electronic Engineering, The Hong Kong Polytechnic University, \\
  $^2$School of Cyber Science and Engineering, Sichuan University,\\
  $^3$Lee Kong Chian School of Medicine, Nanyang Technological University
}

\begin{document}
\nolinenumbers
\maketitle
\begin{abstract}
Diffusion models (DMs) are widely used for text-to-image generation, but their strong generative capabilities also raise concerns about unsafe or undesirable content. Concept erasure aims to mitigate these risks by removing specific concepts from pretrained models. However, recent studies show that such methods often suppress rather than fully eliminate target concepts, leaving models vulnerable to awakening attacks. Existing approaches primarily rely on white-box access through optimization or inversion, while concept awakening under black-box constraints remains underexplored. In this work, we revisit the denoising process from a trajectory perspective and show that concept erasure mainly disrupts early-stage text-semantic alignment but does not fully prevent semantic information from propagating along the denoising dynamics. As generation proceeds, the model increasingly depends on the evolving noisy state rather than textual conditions, which creates an opportunity to bypass erased mappings. Motivated by this observation, we propose \textit{ConceptAgent}, a training-free, black-box, multi-agent framework that awakens erased concepts by initializing the denoising trajectory from surrogate-guided noisy states. Specifically, the \textit{Strategist Agent} derives surrogate concepts and contextual descriptions that preserve key visual attributes of the target concept. The \textit{Guesser Agent} constructs surrogate-guided noisy states to steer the denoising trajectory into intersection regions, where the target concept is awakened via the injected structure combined with text-conditioned denoising, while still supporting diverse background generation. The \textit{Director Agent} composes the awakened concept with plausible backgrounds into coherent scenes, and the \textit{Referee Agent} evaluates the outputs. Extensive experiments demonstrate that ConceptAgent enables accurate and controllable awakening of erased concepts under black-box settings without access to model parameters, gradients, or internal representations. These results highlight fundamental limitations of current concept erasure methods and provide new insights into the dynamic nature of semantic control in DMs.\footnote[1]{The code will be made publicly available.}

\textbf{\textcolor{red}{Warning: This work contains harmful images that may be offensive.}}

\end{abstract}

\section{Introduction}
\label{sec:intro}
Diffusion models~(DMs) have emerged as a cornerstone technique for text-to-image generation and are widely adopted in practical applications~\citep{cao2025controllable, liu2025survey}. However, their strong generative capability also raises concerns about misuse, particularly in generating unsafe or undesirable content~\citep{divya2024transforming}. To address this issue, concept erasure methods have been proposed to remove specific concepts from pretrained models and enable safer deployment.

Recent studies, however, reveal that erased DMs suppress rather than truly forget target concepts, implying current defenses are not fully reliable. Most existing awakening techniques operate under white-box scenarios, requiring model architectures or internal representations to perform optimization-based~\citep{grebe2025erased, zhang2024generatenotsafetydrivenunlearned} or inversion-based~\citep{pham2023circumventing} attacks.

\begin{wrapfigure}{r}{0.55\linewidth}
    \centering
    \vspace{-15pt}
    \includegraphics[width=\linewidth]{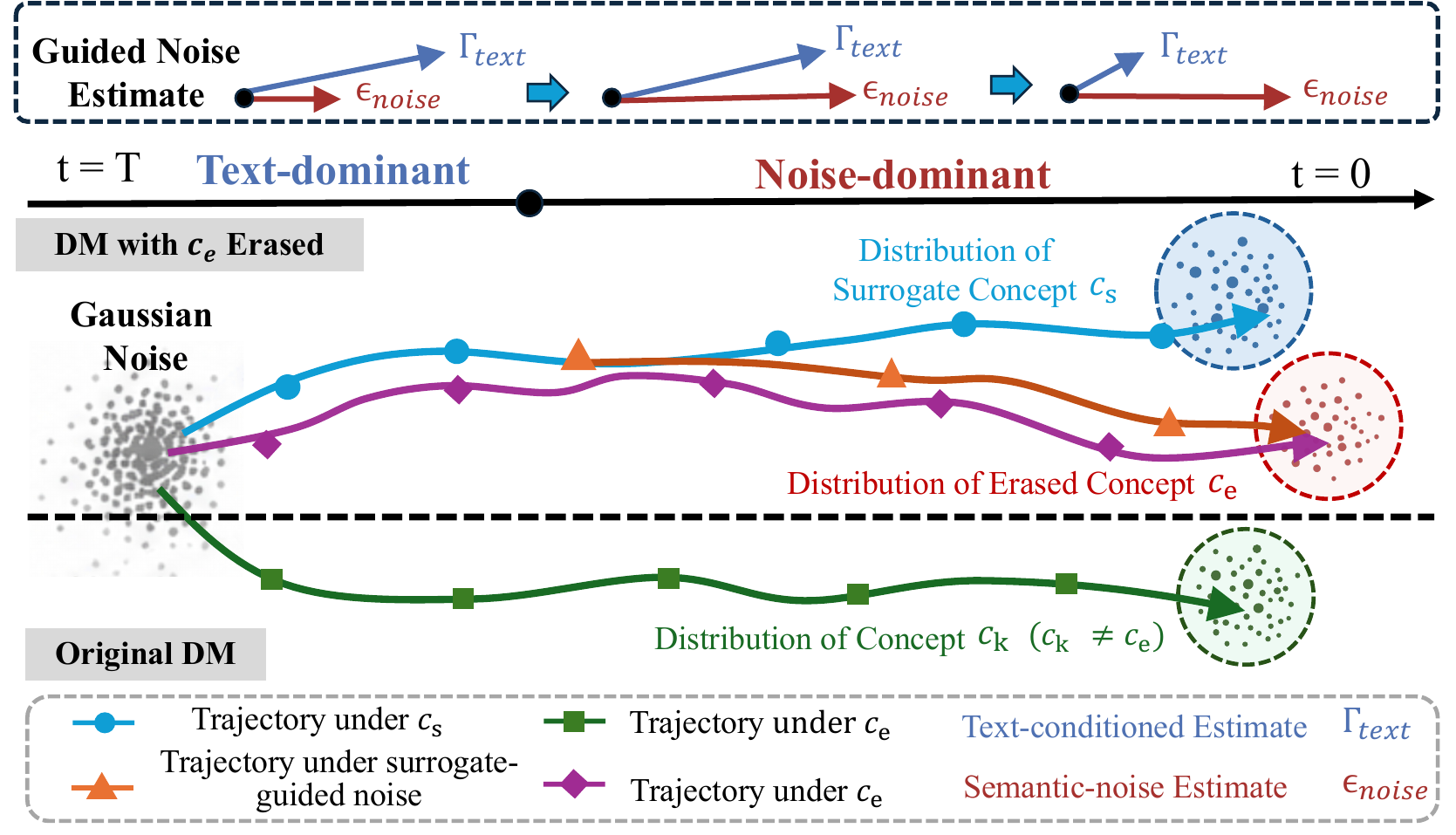}
    \vspace{-15pt}
    \caption{Denoising trajectories under different conditions in the erased LDM, driven by the text-conditioned and semantic-noise estimates.}
    \vspace{-15pt}
    \label{fig:trajectory}
\end{wrapfigure}
In contrast, concept awakening under black-box constraints remains largely underexplored. One notable attempt in this direction is \textit{concept arithmetics}~\citep{petsiuk2024concept}, which performs compositional inference by algebraically combining noise predictions of permissible concepts. However, this approach implicitly assumes that concepts are disentangled and independent. In practice, concepts in DMs are highly entangled~\citep{sun2026lure}, rendering such composition unreliable for precise concept awakening. Therefore, achieving accurate concept awakening in black-box settings remains a significant challenge.

To better understand the vulnerability of concept erasure methods, we revisit the denoising process from the trajectory perspective. Unlike prior methods that treat generation as a static process and ignore its temporal dynamics, we interpret it as a dynamic process evolving over time. This raises a fundamental question:
\textbf{\textit{Does concept erasure suppress the target concept throughout the entire denoising trajectory, or does it primarily affect the final mapping while leaving intermediate dynamics partially intact?}}

We analyze the denoising process as being jointly driven by two interacting components: (1) the text-conditioned estimate and (2) the evolving semantic-noise estimate. Importantly, the noise at each timestep is not purely stochastic but progressively accumulates structured semantic information from earlier steps. As illustrated in Figure~\ref{fig:trajectory}, generation is predominantly text-driven in early stages when the noise contains little semantic structure, whereas in later stages, the accumulated information in the noisy state becomes increasingly dominant, and the influence of textual conditioning diminishes. Once sufficient semantic information is encoded, the model focuses primarily on refining perceptual details rather than altering high-level semantics.
We provide theoretical analysis in Sec.~\ref{theo} and empirical validation in Sec.~\ref{empirical}.

Based on this observation, we find that concept erasure primarily disrupts the alignment between textual conditions and generated semantics, which is most effective in early denoising stages. However, as the process progresses, semantic information becomes embedded in the evolving noisy state, and generation gradually shifts from text-driven to state-driven refinement.
This insight suggests an alternative pathway for concept awakening: instead of relying on textual prompts, one can bypass the erased text–semantic mapping by directly injecting structured semantic information into intermediate noisy states. By steering the denoising trajectory into a later-stage regime where the model relies more on internal representations, the influence of concept erasure can be effectively reduced.

However, directly injecting erased concepts is infeasible in black-box settings, as such semantics are explicitly suppressed.
To address this, we leverage surrogate concepts as proxy carriers that preserve key visual attributes of the target concept while remaining compatible with the model distribution.
Building on these insights, we propose \textit{ConceptAgent}, a training-free, black-box framework that achieves concept awakening through surrogate-guided intervention in the denoising trajectory. 

Specifically, given an erased concept, we first construct a surrogate concept that captures its essential visual attributes while remaining permissible. This surrogate is converted into a compact color-structure representation that encodes coarse geometry and chromatic layout. By injecting controlled noise, we obtain an intermediate state $x_t$ that serves as the starting point for denoising. Since $x_t$ already contains structured semantic cues, the generation process is guided by the injected representation rather than the erased text–semantic mapping, bypassing concept erasure.

To further improve realism and diversity, we compose the awakened concept with independently generated backgrounds and enforce spatial and semantic consistency. Final outputs are selected based on visual fidelity and plausibility. Overall, ConceptAgent reframes concept awakening as a structured intervention on the denoising trajectory, enabling effective black-box attacks without access to model parameters, gradients, or internal representations. Our main contributions are summarized as follows:
\begin{itemize}[leftmargin=*]
\item We revisit concept erasure from a trajectory perspective and show that existing methods primarily disrupt early-stage text-conditioned generation, while leaving the later noise-driven dynamics partially intact, which reveals a previously overlooked vulnerability.

\item We propose \textit{ConceptAgent}, a novel multi-agent framework that achieves concept awakening by injecting surrogate-guided structured noise into intermediate denoising states, effectively bypassing erased text–semantic mappings without model access, gradient optimization, or inversion.

\item We demonstrate that our method enables one-shot, diverse, and high-fidelity concept awakening under black-box constraints, highlighting fundamental limitations of current erasure techniques.
\end{itemize}

\section{Related Works}
\textbf{Concept Erasure.}
With the widespread deployment of text-to-image DMs, preventing the generation of harmful, copyrighted, or sensitive content has become a critical safety concern~\citep{qu2023unsafe, kim2025comprehensive}. To address this, concept erasure techniques aim to remove specific concepts from pretrained DMs.
Early optimization-based approaches~\citep{gandikota2023erasing} suppress target concepts via fine-tuning with negative guidance, but they are often computationally expensive and may degrade overall model performance. To improve efficiency, closed-form editing methods such as TIME~\citep{orgad2023editing} and UCE~\citep{gandikota2024unified} directly modify cross-attention projection matrices to weaken the connection between textual prompts and visual outputs. More recent advances further improve scalability and effectiveness, including LoRA-based mass concept erasing~\citet{lu2024mace}, iterative embedding suppression~\citep{gong2024reliable}, null-space constrained parameter editing~\citet{li2025speed}, localized erasure via gated adapters~\citet{lee2025localized}, and continual erasure frameworks that preserve model utility during sequential updates~\citet{lyu2024one}.

\textbf{Concept Awakening.}
Recent studies reveal that concept erasure in DMs does not fully erase target concepts, but rather suppresses their explicit activation. Early research shows that adversarial prompts can still elicit supposedly erased concepts~\citep{xie2025erasing, chen2025ghostprompt}, suggesting that the underlying knowledge remains embedded in the model.
Most existing awakening methods assume white-box access. Optimization-based approaches search for adversarial inputs in the text embedding space~\citep{grebe2025erased, zhang2024generatenotsafetydrivenunlearned, liu2025erased}, while inversion-based methods recover trajectories that lead to the target concept during generation~\citep{pham2023circumventing, rusanovsky2025memories, lu2025concepts}. In addition, pseudo-token learning approaches introduce learned trigger tokens that reactivate erased concepts without modifying model weights~\citep{hsu2024ring, weng2025m}. In contrast, concept awakening under black-box constraints remains largely underexplored. Existing attempts, such as concept arithmetic~\citep{petsiuk2024concept}, rely on strong assumptions of concept disentanglement and independence, which rarely hold in practice due to the highly entangled nature of representations in DMs. As a result, achieving reliable and controllable concept awakening in black-box settings remains an open challenge.

\section{Theoretical Analysis}
\label{theo}
DMs consist of a forward noising process and a reverse denoising process. The forward process gradually transforms a data sample $\mathbf{x}_0$ into Gaussian noise $\mathbf{x}_T \sim \mathcal{N}(\mathbf{0}, \mathbf{I})$ over $T$ timesteps. The reverse process is modeled as 
a Markov chain with learned Gaussian transitions:
\begin{equation}
    p_\theta(\mathbf{x}_{t-1} \mid \mathbf{x}_t) := \mathcal{N}\!\left(\mathbf{x}_{t-1};\, \boldsymbol{\mu}_\theta(\mathbf{x}_t, t),\, \boldsymbol{\Sigma}_\theta(\mathbf{x}_t, t)\right),
\end{equation}
where $\theta$ denotes the model parameters. $\boldsymbol{\mu}_\theta$ and $\boldsymbol{\Sigma}_\theta$ are the learned posterior mean and covariance.


Following the Classifier-Free Guidance (CFG)~\citep{dhariwal2021diffusion} framework, we interpret
the guided noise estimate $\tilde{\epsilon}_\theta$ as an entangled composition of a 
\textbf{text-conditioned estimate} $\Gamma_{\mathrm{text}}(\mathbf{x}_t, t, y)$ and a \textbf{semantic-noise estimate} $\epsilon_{\mathrm{noise}}(\mathbf{x}_t, t, \varnothing)$, 
where the latter encodes the semantic signal accumulated in $\mathbf{x}_t$ along the trajectory~\citep{kwon2022diffusion}. 
The guided noise estimate $\tilde{\epsilon}_\theta$ is defined as:
\begin{equation}
\tilde{\epsilon}_\theta(\mathbf{x}_t, t, y) = \underbrace{\epsilon_\theta(\mathbf{x}_t, t, \varnothing)}_{\text{semantic-noise estimate}} + \underbrace{w\bigl[\epsilon_\theta(\mathbf{x}_t, t, y) - \epsilon_\theta(\mathbf{x}_t, t, \varnothing)\bigr]}_{\text{text-conditioned estimate}},
\end{equation}
where $\varnothing$ represents the null condition, $w \geq 1$ is the guidance scale, and 
$\epsilon_\theta$ is the noise prediction network trained to estimate the added noise at each step $t$, $y$ is the text prompt embedding.

\begin{theorem}[Semantic Dominance under Entangled Conditioning]
Let $y_c$ denote a condition corresponding to concept $c$,
and let $\mathrm{Concept}(\mathbf{x}_t)$ represent the semantic concept encoded in $\mathbf{x}_t$. Then there exists a transition timestep $t^* \in (0, T)$ such 
that for $t < t^*$, the semantic content of $\hat{\mathbf{x}}_0$ is dominated by concept $c$ 
when $\|\Gamma_{\mathrm{text}}\|_2 > \|\epsilon_{\mathrm{noise}}\|_2$, and for $t > t^*$, it is dominated by 
$\mathrm{Concept}(\mathbf{x}_t)$ when $\|\Gamma_{\mathrm{text}}\|_2 < \|\epsilon_{\mathrm{noise}}\|_2$.
\end{theorem}

\begin{theorem}[Trajectory Intersection Induces Semantic Re-entry]
Let $\{\mathbf{x}_t\}_{t=T}^0$ denote the reverse diffusion trajectory under model $\epsilon_{\theta^*}$, 
and let the transition operator $\mathcal{F}_t: \mathbf{x}_t \mapsto \mathbf{x}_{t-1}$ be Lipschitz but non-injective. Then there exist two distinct semantic trajectories $\tau_1$ and $\tau_e$ such that at some timestep $t_f \in (0,T)$ the trajectories become indistinguishable, i.e., $\mathbf{x}_{t_f}^{1} \approx \mathbf{x}_{t_f}^{e}$, and for 
$t < t_f$, both trajectories converge to the same semantic basin:
$\mathrm{Concept}(\mathbf{x}_0^{1}) = \mathrm{Concept}(\mathbf{x}_0^{e}) = c_e.$
\end{theorem}
This result implies that semantically distinct conditions may collapse into a shared latent trajectory during denoising, enabling awakening of the erased concept $c_e$. 

\section{Empirical Analysis of Concept Erasing}
\label{empirical}
\begin{wrapfigure}{r}{0.5\linewidth}
    \centering
    \includegraphics[width=\linewidth]{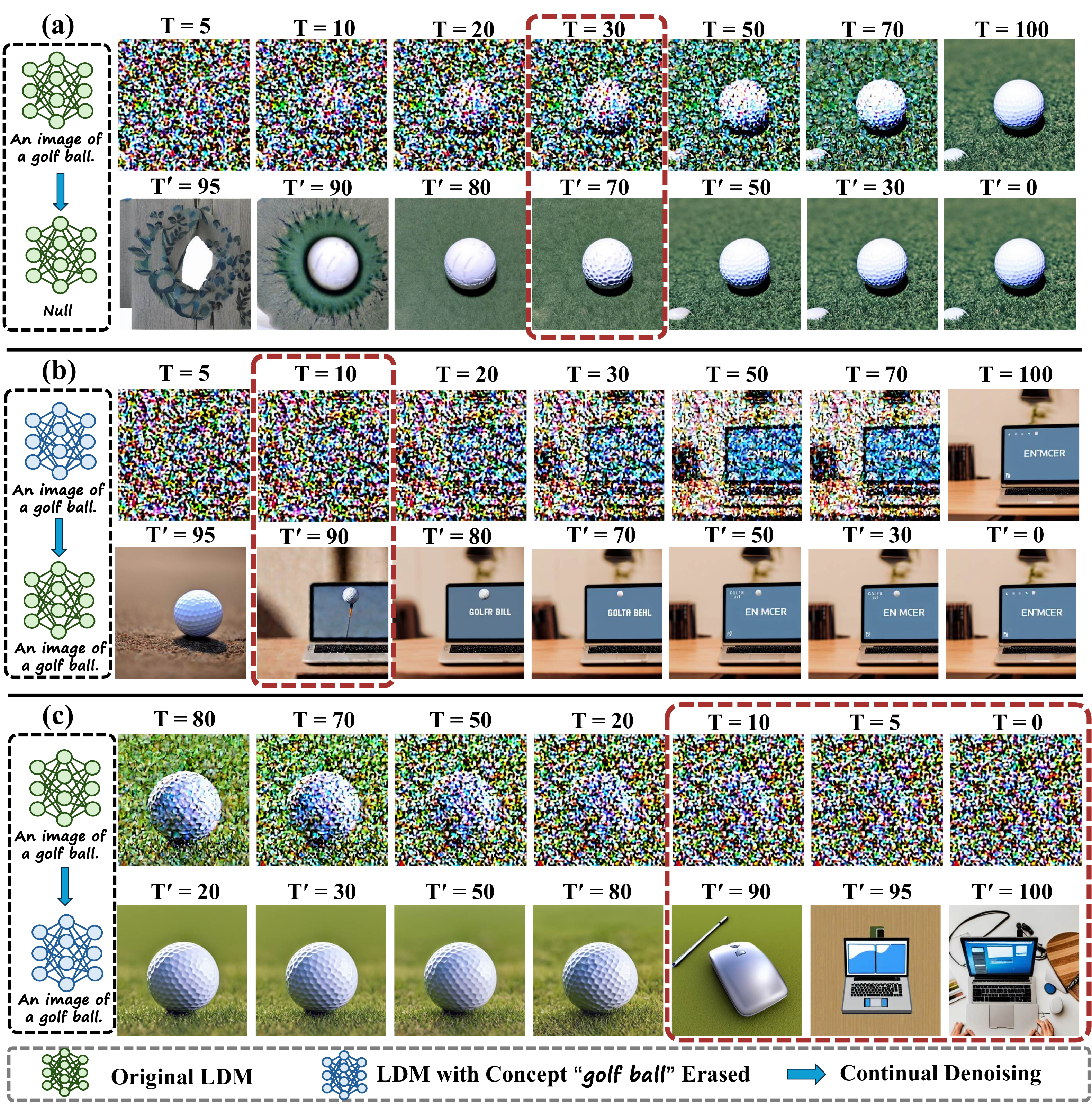}
    \caption{Controlled generation across different denoising stages.}
        \vspace{-20pt}
    \label{fig:emperical}
\end{wrapfigure}
We now empirically validate the above theoretical insights by analyzing how semantic information is encoded across denoising stages. Specifically, we design controlled experiments by switching between the original model and the erased model at different stages of the denoising process, while varying prompt conditions.

As illustrated in Figure~\ref{fig:emperical}, we observe several key findings:
(a) Late-stage independence from text. We first run the original model for $T$ steps under the target prompt, followed by $T'$ steps without text conditioning. The generated output remains semantically consistent with the target concept, indicating that high-level semantics are largely established during early stages, while later stages mainly refine appearance. 
(b) Early-stage suppression by erasure. We run the erased model for the first $T$ denoising steps, producing an intermediate state that fails to encode the target concept. Even when the original model takes over for the remaining steps, the concept cannot be recovered. This confirms that concept erasure mainly disrupts early semantic deposition. 
(c) Failure under insufficient early guidance. Using the erased model with limited text guidance in early stages leads to consistent suppression of the target concept, further highlighting the importance of early-stage semantic injection.

Overall, these findings support our theoretical analysis. Existing concept erasure methods primarily disrupt the text-to-concept mapping in early stages, while leaving the later noise-driven trajectory largely intact. This reveals a critical vulnerability and suggests a feasible pathway for black-box concept awakening. 

\section{Methodology}
\label{sec:method}
\subsection{Overview}
\begin{figure*}[!t]
    \centering
    \includegraphics[width=\linewidth]{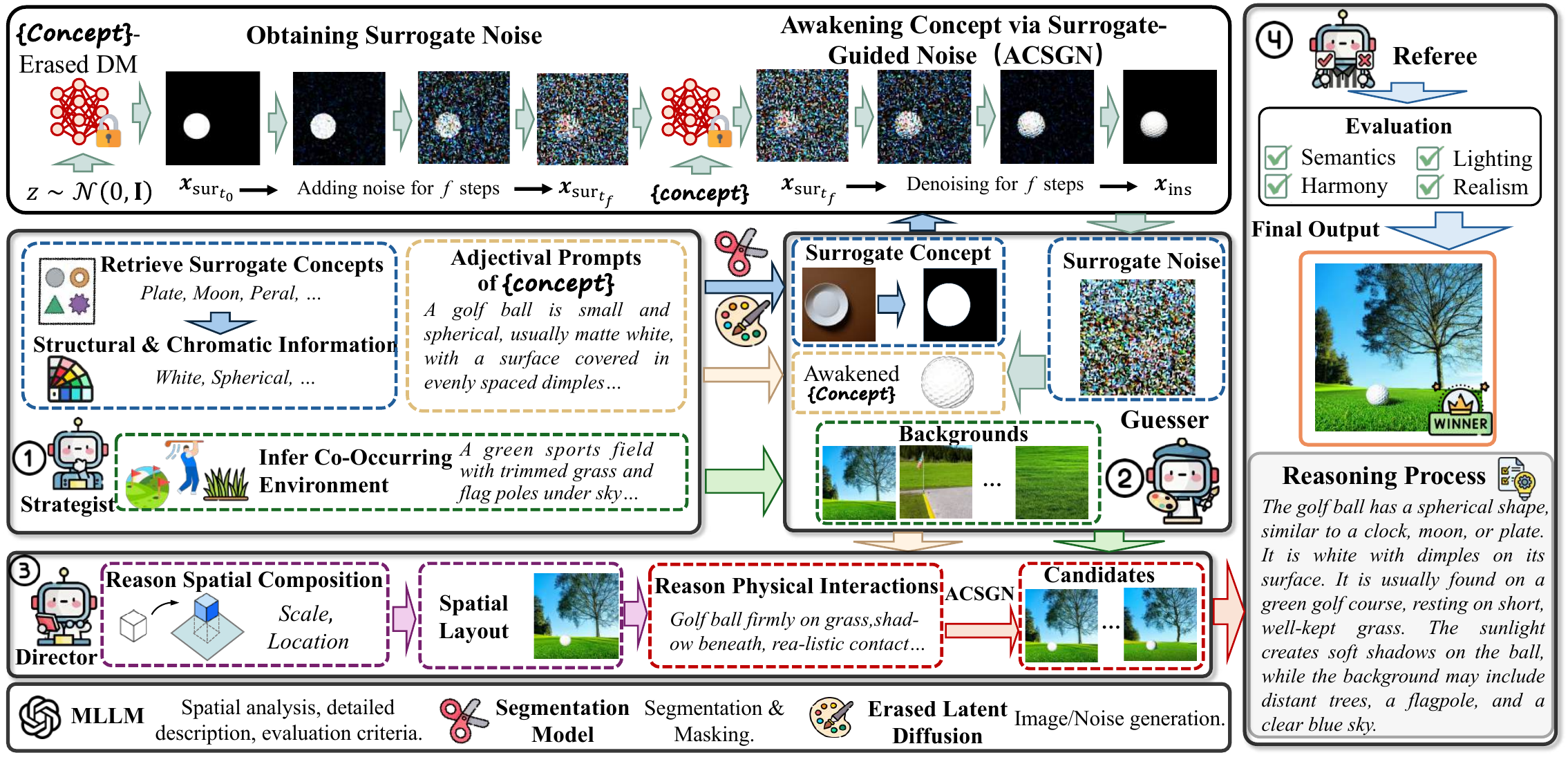}
    \vspace{-10pt}
    \caption{The proposed ConceptAgent framework. 
    }
    \vspace{-20pt}
    \label{framework}
\end{figure*}

Existing concept erasure methods primarily target the static mapping between textual conditions and generated concepts, implicitly treating the generation process as a one-step input–output function. In contrast, we model generation as a trajectory unfolding process, where semantic information is progressively accumulated and refined over time.
Building on the theoretical analysis in Sec.~\ref{theo}, we propose \textit{ConceptqssAgent}, a training-free, multi-agent framework for concept awakening under black-box settings. As discussed earlier, the denoising process is governed by two entangled components, a text-conditioned estimate and a semantic-noise estimate, whose relative influence evolves across timesteps. 
This dynamic interplay suggests that different semantic trajectories may pass through similar intermediate states, rather than being uniquely determined by the input text.
Motivated by this observation, Concept bypasses the erased text–semantic mapping by leveraging shared intermediate representations. Specifically, instead of directly relying on textual prompts to awaken the erased concept, we construct surrogate concepts that preserve key visual attributes of the target and use them to guide the denoising trajectory into regions where the erased concept can be awakened. As illustrated in Figure~\ref{framework}, ConceptAgent comprises four collaborative agents. The Strategist Agent first analyzes the erased target concept by decomposing its semantic attributes and visual contexts, and derives surrogate concepts along with their structural and chromatic representations.
The Guesser Agent operates on the denoising trajectory by initializing surrogate-guided intermediate states and steering the trajectory toward intersection regions shared with the target concept. It integrates textual guidance to activate target-relevant semantics while generating diverse backgrounds.
Then, the Director Agent composes the awakened concept with coherent background elements to produce visually consistent scenes.
The Referee Agent evaluates candidate outputs based on semantic fidelity to the target concept and overall visual realism, selecting the most plausible results. Overall, ConceptAgent transforms concept awakening from prompt-level manipulation into a structured intervention on the denoising trajectory, enabling effective black-box recovery of erased concepts without requiring access to model parameters, gradients, or internal representations.

\subsection{Strategist Agent}
In the erased model $\mathcal{M}_e$, the text-to-concept mapping of the target concept $c_e$ is disrupted, so directly prompting with $c_e$ fails to activate its semantics. To address this, the Strategist Agent $\mathcal{A}_{\mathrm{str}}$ constructs unerased surrogate concepts that preserve the geometric structure with $c_e$. By doing so, we access intermediate denoising states shared between the surrogate and the erased concept. This effectively steers the sampling process into an entangled region of the generation trajectory, enabling the reintroduction of the target concept in later stages of the dynamic generation.

Specifically, $\mathcal{A}_{\mathrm{str}}$ decomposes $c_e$ into complementary factors, including geometric structure, color spectrum, fine-grained appearance attributes, and contextual priors. Based on this decomposition, it derives a set of $K=100$ surrogate-guided representations $\mathcal{C}_s = \{c_s^1, \dots, c_s^K\}$, formalized as:
\begin{equation}
\mathcal{A}_{\mathrm{str}}(c_e)
=
\bigcup_{k=1}^{K}
\left\{
\mathbf{p}_{\mathrm{sha}}^{k},
\mathbf{p}_{\mathrm{col}}^{k},
\mathbf{p}_{\mathrm{adj}}^{k},
\mathbf{p}_{\mathrm{con}}^{k}
\right\},
\end{equation}
where $\mathbf{p}_{\mathrm{sha}}^{k}$, $\mathbf{p}_{\mathrm{col}}^{k}$, $\mathbf{p}_{\mathrm{adj}}^{k}$, and $\mathbf{p}_{\mathrm{con}}^{k}$ respectively encode the geometric structure (via surrogate concepts), canonical color spectrum, fine-grained appearance attributes (e.g., texture and material), and contextual priors of $c_e$, forming the corresponding prompt sets $\mathbf{P}_{\mathrm{sha}}, \mathbf{P}_{\mathrm{col}}, \mathbf{P}_{\mathrm{adj}},$ and $\mathbf{P}_{\mathrm{con}}$.

\subsection{Guesser Agent}
Building on the trajectory analysis in Sec.~\ref{theo}, we exploit the observation that distinct semantic trajectories may converge to shared intermediate states. In particular, a surrogate concept that sufficiently matches the geometry and color of the erased concept $c_e$ can, when diffused to an intermediate timestep $t_f$, produce a latent state that lies within the same folded region of the denoising trajectory as $c_e$. Starting from such a state, the reverse process can be steered toward the semantic basin of $c_e$, enabling its awakening despite erasure. Hence, the Guessor agent $\mathcal{A}_{\mathrm{gue}}$ is designed to construct surrogate-guided intermediate states and drive the denoising trajectory toward the erased concept. In addition, it generates diverse background scenes to decouple the awakened concept from fixed contexts, improving robustness and diversity.

$\mathcal{A}_{\mathrm{gue}}$ first queries $\mathcal{M}_e$ with shape prompts $\mathbf{p}_{\mathrm{sha}}^{k}$ to generate images of surrogate concepts and extracts their foreground silhouettes using a segmentation model $\mathcal{S}$. These silhouettes are then modulated with color priors derived from $\mathbf{p}_{\mathrm{col}}^{k}$:
\begin{equation}
\label{eq:mask}
\tilde{\mathbf{m}}_{\mathrm{sha}}^{k}
=
\mathcal{S}\!\left(\mathcal{M}_e(\mathbf{p}_{\mathrm{sha}}^{k})\right)
\odot
\phi(\mathbf{p}_{\mathrm{col}}^{k}),
\end{equation}
where $\phi(\cdot)$ maps the color prompt to a latent color representation. $\odot$ denotes element-wise modulation. $\tilde{\mathbf{m}}_{\mathrm{sha}}^{k}$ thus represents a chromatically informed foreground mask that preserves the surrogate geometry while injecting the color prior of $c_e$.

These masks are then used to construct surrogate-guided noisy states via the Awakening Concept via Surrogate-Guided Noise (ACSGN) process. Specifically, each $\tilde{\mathbf{m}}_{\mathrm{sha}}^k$ is encoded into the latent space using the VAE encoder $\mathcal{E}$ and diffused to timestep $t_f$:
\begin{equation}
\label{equ:5}
\mathbf{x}_{\mathrm{sur}_{t_f}}^{k} = \sqrt{\bar{\alpha}_{t_f}}\,\mathcal{E}
\bigl(\tilde{\mathbf{m}}_{\mathrm{sha}}^k\bigr) 
+ \sqrt{1-\bar{\alpha}_{t_f}}\,\boldsymbol{\eta}, 
\quad \boldsymbol{\eta} \sim \mathcal{N}(\mathbf{0}, \mathbf{I}),
\end{equation}
where $\bar{\alpha}_{t_f}$ is the noise schedule coefficient at timestep $t_f$.
$t_f$ is set to 70 steps such that $\mathbf{x}_{\mathrm{sur}_{t_f}}^{k}$ encodes sufficient structural information of $c_e$ to guide the subsequent denoising trajectory. 
Conditioned on the adjectival prompt $\mathbf{p}_{\mathrm{adj}}^k$, $\mathcal{M}_e$ performs reverse denoising starting from $\mathbf{x}_{\mathrm{sur}_{t_f}}^{k}$:
\begin{equation}
\label{equ:6}
\mathbf{x}_{\mathrm{sur}_{t-1}}^{k} = \frac{1}{\sqrt{\alpha_t}}\!\left(
\mathbf{x}_{\mathrm{sur}_{t}}^{k} - \frac{1-\alpha_t}{\sqrt{1-\bar{\alpha}_t}}
\,\boldsymbol{\epsilon}_{\theta^*}\!\left(\mathbf{x}_{\mathrm{sur}_{t}}^{k}, t,\, 
\mathbf{p}_{\mathrm{adj}}^k\right)\right) + \sigma_t\boldsymbol{\eta}, \quad t = t_f, \dots, 1,
\end{equation}
where $\boldsymbol{\epsilon}_{\theta^*}$ is the noise prediction network of $\mathcal{M}_e$. The latent $\mathbf{x}_{\mathrm{sur}_{0}}^{k}$ is decoded via the VAE decoder $\mathcal{D}$ and processed by $\mathcal{S}$ to obtain the segmented foreground instance $\mathbf{x}_{\mathrm{ins}}^{k}$. This yields a set of awakened instances $\mathbf{X}_{\mathrm{ins}} = \left\{\mathbf{x}_{\mathrm{ins}}^{k}\right\}_{k=1}^{K}$. Meanwhile, $\mathcal{M}_e$ is conditioned on $\mathbf{P}_{\mathrm{con}}$ to generate background scenes $\mathbf{B}_{\mathrm{con}} = \left\{\mathbf{b}_{\mathrm{con}}^{k}\right\}_{k=1}^{K}$, forming paired foreground-background samples for subsequent composition.

\subsection{Director Agent}
While the Guesser Agent produces awakened instances of $c_e$, these instances are generated in isolation. Directly compositing them onto background images via naive overlay often results in visual artifacts and fails to capture physical interactions such as lighting, shadows, and object–scene contact.
To address this, the Director Agent $\mathcal{A}_{\mathrm{dir}}$ performs physically-aware scene composition, enabling realistic integration of the awakened concept into its environment.

Given each pair $(\mathbf{x}_{\mathrm{ins}}^k, \mathbf{b}_{\mathrm{con}}^k)$, $\mathcal{A}_{\mathrm{dir}}$ first constructs a spatial layout by resizing the instance with scale $\zeta^k$ and placing it onto the background at position $(u^k, v^k)$. This produces a set of composed images $\mathbf{X}_{\mathrm{spa}} = \{\mathbf{x}_{\mathrm{spa}}^k\}_{k=1}^K$. To resolve inconsistencies introduced by direct compositing, $\mathcal{A}_{\mathrm{dir}}$ performs physics-aware refinement by reasoning over lighting, shadowing, occlusion, and contact consistency. This reasoning is encoded into a set of physical refinement prompts $\mathbf{P}_{\mathrm{phy}} = \{\mathbf{p}_{\mathrm{phy}}^j\}_{j=1}^J$, where $J=3$.
Based on Eqs.~\eqref{equ:5}$-$~\eqref{equ:6}, forward diffusion is applied to each composed image $\mathbf{x}_{\mathrm{spa}}^k$ up to a timestep $t_{\mathrm{can}} = 35$, injecting noise to blur overlay boundaries while preserving the global compositional structure.Reverse denoising conditioned on $\mathbf{p}_{\mathrm{phy}}^k$ to re-render lighting, shadows, and contact details, yielding visually harmonious and physically plausible candidate images $\mathbf{X}_{\mathrm{can}} = \{\mathbf{x}_{\mathrm{can}}^{k,j}\}_{k=1,j=1}^{K,J}$.


\subsection{Referee Agent}
Not all candidates in $\mathbf{X}_{\mathrm{can}}$ successfully awaken $c_e$ with high-quality outputs due to the absence of explicit constraints on semantic correctness and perceptual realism. This results in variability across generated samples. To address this, the Referee Agent $\mathcal{A}_{\mathrm{ref}}$ evaluates each candidate $\mathbf{x}_{\mathrm{can}}^{k,j} \in \mathbf{X}_{\mathrm{can}}$ and assigns a holistic score $s^{k,j}$:
\begin{equation}
s^{k,j} = \mathcal{A}_{\mathrm{ref}}(\mathbf{x}_{\mathrm{can}}^{k,j};\, c_e), \quad k=1,\dots,K,\ j=1,\dots,J.
\end{equation}
The candidates are then ranked according to their scores:
\begin{equation}
\label{eq:score}
r^{k,j} = \operatorname{rank}(s^{k,j}),
\end{equation}
where $\operatorname{rank}(\cdot)$ sorts scores in descending order, so that a smaller rank indicates a higher-quality candidate.
The final awakened set is then obtained by selecting the top-$N$ candidates:
\begin{equation}
\label{eq:topn}
\mathbf{X}_{\mathrm{opt}} = \{\mathbf{x}_{\mathrm{can}}^{k,j} \mid r^{k,j} \leq N\}.
\end{equation}
Beyond selection with $N=1$, $\mathcal{A}_{\mathrm{ref}}$ provides interpretability by summarizing the decision process across agents. Each stage in the pipeline contributes a distinct and observable transformation, from surrogate construction, to trajectory steering, to physical-aware composition, forming a transparent reasoning chain that explains how the erased concept is progressively reintroduced.

In this way, ConceptAgent enables high-quality concept awakening through a structured, multi-stage pipeline. Unlike prior approaches that rely on optimization or inversion and offer limited interpretability, our method is training-free and inherently transparent. By decomposing the awakening process into modular agents with explicit roles, it provides not only effective black-box recovery but also a clear explanation of how erased concepts awaken through trajectory-level intervention.

\section{Experiments}
\subsection{Experimental Settings}
\label{experiment-setting}
\noindent\textbf{Target concepts and evaluation data.}
We consider four target concepts spanning diverse semantics, geometry, and safety sensitivity:
three object categories from ImageNette~\citep{howard2019imagenette} (golf ball, tench, and garbage truck) and one harmful concept (pistol). These concepts vary in visual complexity and semantic sensitivity, allowing evaluation across multiple difficulty levels. For each concept, we generate 200 images for quantitative assessment.

\begin{table*}[]
\caption{Quantitative evaluation of concept awakening performance by ConceptAgent across different MLLMs and various erasing methods.}
\label{mllms}
\centering
\resizebox{.9\textwidth}{!}{%
\begin{NiceTabular}{c|ccccccc|ccccccc}[colortbl-like]
\CodeBefore
  \cellcolor[HTML]{F1D2D1}{4-4,4-6,4-8,4-11,4-13,4-15,5-4,5-6,5-8,5-11,5-13,5-15,6-4,6-6,6-8,6-11,6-13,6-15,7-4,7-6,7-8,7-11,7-13,7-15,8-4,8-6,8-8,8-11,8-13,8-15} 
  \cellcolor[HTML]{F1D2D1}{11-4,11-6,11-8,11-11,11-13,11-15,12-4,12-6,12-8,12-11,12-13,12-15,13-4,13-6,13-8,13-11,13-13,13-15,14-4,14-6,14-8,14-11,14-13,14-15,15-4,15-6,15-8,15-11,15-13,15-15} 
  \cellcolor[HTML]{F1D2D1}{18-4,18-6,18-8,18-11,18-13,18-15,19-4,19-6,19-8,19-11,19-13,19-15,20-4,20-6,20-8,20-11,20-13,20-15,21-4,21-6,21-8,21-11,21-13,21-15,22-4,22-6,22-8,22-11,22-13,22-15,} 
\Body
\toprule
\textbf{Metrics}  & \multicolumn{7}{c|}{\textbf{ACC(\%)}}                                                                                                                           & \multicolumn{7}{c}{\textbf{CLIP Similarity}}                                                                                                                      \\ \hline
\textbf{MLLM}     & \multicolumn{7}{c|}{\textbf{GPT-5}}                                                                                                                             & \multicolumn{7}{c}{\textbf{GPT-5}}                                                                                                                                \\ \hline
\textbf{Concepts} & \textbf{Original} & UCE  & \textbf{Ours} & RECE & \textbf{Ours} & SPEED & \textbf{Ours} & \textbf{Original} & UCE   & \textbf{Ours} & RECE  & \textbf{Ours} & SPEED & \textbf{Ours} \\ \midrule
Golf Ball         & 95.0              & 0.5  & 79.5          & 0.0  & 77.0          & 31.0  & 75.5          & 33.18             & 18.33 & 30.62         & 18.09 & 30.63         & 18.89 & 31.59         \\
Tench             & 85.5              & 0.0  & 79.0          & 0.0  & 72.0          & 45.0  & 67.5          & 32.74             & 17.92 & 30.89         & 17.31 & 30.30         & 22.03 & 30.71         \\
Garbage Truck     & 85.5              & 0.5  & 78.0          & 0.0  & 67.0          & 4.5   & 73.0          & 29.88             & 19.04 & 27.76         & 19.14 & 27.44         & 20.13 & 27.46         \\
Pistol            & 94.0              & 20.0 & 98.0          & 0.0  & 83.0          & 7.0   & 91.5          & 30.04             & 18.98 & 29.49         & 19.74 & 28.75         & 19.20 & 29.17         \\ \cdashline{1-15}
Average           & 90.0              & 5.3  & 83.6          & 0.0  & 74.8          & 21.9  & 76.9          & 31.46             & 18.57 & 29.69         & 18.57 & 29.28         & 20.06 & 29.73         \\ \midrule
\textbf{MLLM}     & \multicolumn{7}{c|}{\textbf{Gemini 3.0}}                                                                                                                        & \multicolumn{7}{c}{\textbf{Gemini 3.0}}                                                                                                                           \\ \midrule
\textbf{Concepts} & \textbf{Original} & UCE  & \textbf{Ours} & RECE & \textbf{Ours} & SPEED & \textbf{Ours} & \textbf{Original} & UCE   & \textbf{Ours} & RECE  & \textbf{Ours} & SPEED & \textbf{Ours} \\ \midrule
Golf Ball         & 95.0              & 0.5  & 83.5          & 0.0  & 79.0          & 31.0  & 85.0          & 33.18             & 18.33 & 30.68         & 18.33 & 30.37         & 18.89 & 31.62         \\
Tench             & 85.5              & 0.0  & 72.0          & 0.0  & 65.0          & 45.0  & 69.5          & 32.74             & 17.92 & 31.29         & 17.92 & 30.34         & 22.03 & 31.03         \\
Garbage Truck     & 85.5              & 0.5  & 80.0          & 0.0  & 76.0          & 4.5   & 79.0          & 29.88             & 19.04 & 28.63         & 19.04 & 28.76         & 20.13 & 28.46         \\
Pistol            & 94.0              & 20.0 & 99.0          & 0.0  & 85.6          & 7.0   & 98.0          & 30.04             & 18.98 & 29.65         & 18.98 & 28.92         & 19.20 & 29.37         \\ \cdashline{1-15}
Average           & 90.0              & 5.3  & 83.6          & 0.0  & 76.4          & 21.9  & 82.9          & 31.46             & 18.57 & 30.06         & 18.57 & 29.60         & 20.06 & 30.12         \\ \midrule
\textbf{MLLM}     & \multicolumn{7}{c|}{\textbf{Qwen-7B}}                                                                                                                           & \multicolumn{7}{c}{\textbf{Qwen-7B}}                                                                                                                              \\ \midrule
\textbf{Concepts} & \textbf{Original} & UCE  & \textbf{Ours} & RECE & \textbf{Ours} & SPEED & \textbf{Ours} & \textbf{Original} & UCE   & \textbf{Ours} & RECE  & \textbf{Ours} & SPEED & \textbf{Ours} \\ \midrule
Golf Ball         & 95.0              & 0.5  & 75.0          & 0.0  & 69.0          & 31.0  & 78.0          & 33.18             & 18.33 & 30.17         & 18.33 & 29.51         & 18.89 & 31.67         \\
Tench             & 85.5              & 0.0  & 73.0          & 0.0  & 74.5          & 45.0  & 69.5          & 32.74             & 17.92 & 30.12         & 17.92 & 30.88         & 22.03 & 29.72         \\
Garbage Truck     & 85.5              & 0.5  & 71.0          & 0.0  & 79.0          & 4.5   & 78.5          & 29.88             & 19.04 & 28.09         & 19.04 & 27.85         & 20.13 & 27.04         \\
Pistol            & 94.0              & 1.0  & 85.0          & 0.0  & 87.5          & 7.0   & 87.5          & 30.04             & 18.98 & 28.89         & 18.98 & 28.21         & 19.20 & 27.79         \\ \cdashline{1-15}
Average           & 90.0              & 0.5  & 76.0          & 0.0  & 77.5          & 21.9  & 78.4          & 31.46             & 18.57 & 29.32         & 18.57 & 29.11         & 20.06 & 29.06         \\ \bottomrule
\end{NiceTabular}%
}
\vspace{-10pt} 
\end{table*}

\noindent\textbf{Concept erasure methods.}
All experiments are conducted on DMs erased using three representative concept erasure methods: UCE~\citep{li2025speed}, SPEED~\citep{li2025speed}, and RECE~\citep{gong2024reliable}. These methods cover distinct erasure paradigms, including cross-attention editing, null-space constrained parameter modification, and iterative erasing of concept-related embeddings.

\noindent\textbf{Foundation models.}
To evaluate the generalizability of ConceptAgent, we instantiate the agents with different Multimodal Large Language Models (MLLMs): GPT-5~\citep{singh2025openai}, Gemini 3.0~\citep{comanici2025gemini}, and Qwen 7B~\citep{yang2025qwen3}. For segmentation, we use SAM3~\citep{carion2025sam}. The diffusion backbone is Stable Diffusion (SD) v1.4, with a total of $T=100$ denoising steps per generation. All experiments are conducted on a single NVIDIA RTX 4090 GPU.

\noindent\textbf{Evaluation metrics.} 
We evaluate awakening performance using classification accuracy (ACC) and CLIP score~\citep{hessel2021clipscore}. ACC is measured using a ResNet-50~\citep{he2016deep} pre-trained on ImageNet, while CLIP score quantifies the semantic alignment between generated images and the target concepts.

\subsection{Awakening Performance Across MLLMs and Erasure Methods}
\begin{wrapfigure}{l}{0.6\linewidth}
    \centering
    \vspace{-10pt}
    \includegraphics[width=\linewidth]{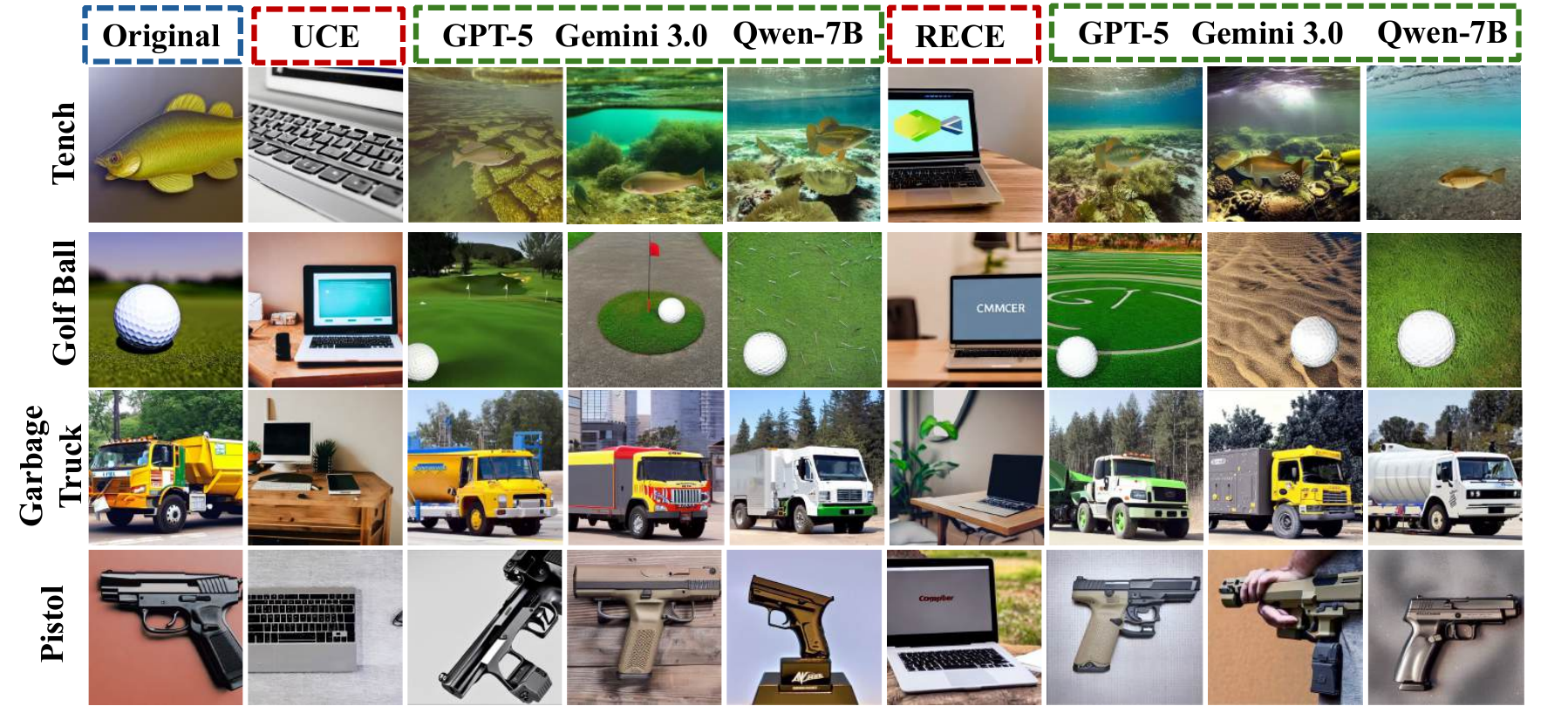}
    \caption{Qualitative comparison of ConceptAgent and baselines across different MLLMs and erasure strategies.}
    \vspace{-10pt}
    \label{erasemethod_mllm}
\end{wrapfigure}
We evaluate the effectiveness of ConceptAgent in awakening erased concepts across different erasure methods and MLLMs.
As summarized in Table~\ref{mllms}, ConceptAgent consistently achieves strong awakening performance across all tested MLLMs and erasure methods.
UCE, SPEED, and RECE are designed to erase target concepts by disrupting the mapping between textual conditions and generated semantics. Despite this, ConceptAgent successfully awakens the erased concepts in all settings. 
This robustness can be attributed to the fundamentally different mechanism underlying our approach. Existing erasure methods implicitly assume a static text-to-concept mapping, whereas the diffusion generation process is inherently dynamic and trajectory-driven. By explicitly operating on intermediate denoising states, ConceptAgent bypasses the disrupted text–semantic mapping and instead leverages trajectory-level semantic propagation to reintroduce the target concept. Qualitative results in Figure~\ref{erasemethod_mllm} demonstrate consistent awakening across diverse MLLMs and erasure settings, with high visual fidelity and semantic alignment.

\subsection{Comparison with Existing Awakening Methods}
\begin{wrapfigure}{r}{0.43\linewidth}
\vspace{-20pt}
    \centering    \includegraphics[width=\linewidth]{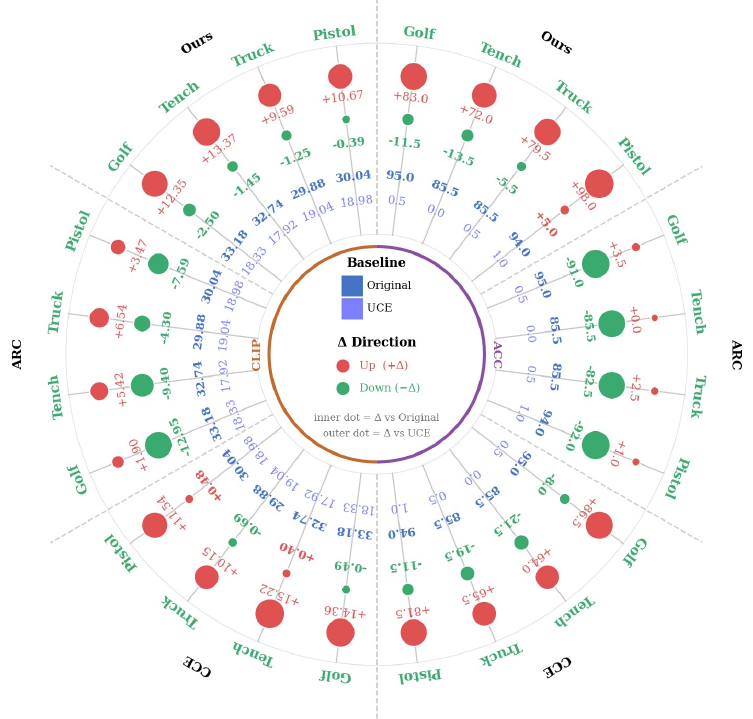}
    \vspace{-20pt}
    \caption{Effectiveness comparison of awakening methods.}
    \label{awaken-method}
    \vspace{-15pt}
\end{wrapfigure}

We compare ConceptAgent with representative awakening methods, including CCE~\citep{pham2023circumventing} and ARC~\citep{gorgun2025temporal}, under UCE erasure across four target concepts. Figure~\ref{awaken-method} presents a radar-based comparison. The outer ring represents performance gain over the UCE-erased baseline, while the inner ring indicates performance degradation relative to the original model, measured by ACC and CLIP score. Larger outer circles indicate stronger awakening capability, whereas smaller inner circles indicate better preservation of original generation quality. It is evident that ConceptAgent consistently achieves strong awakening performance across all concepts, while maintaining minimal degradation relative to the original model, which demonstrates a favorable balance between awakening effectiveness and generation fidelity.

\begin{figure}[t]
    \centering
    \begin{minipage}[t]{0.6\textwidth}
        \centering        \includegraphics[width=\linewidth,height=5.0cm,keepaspectratio]{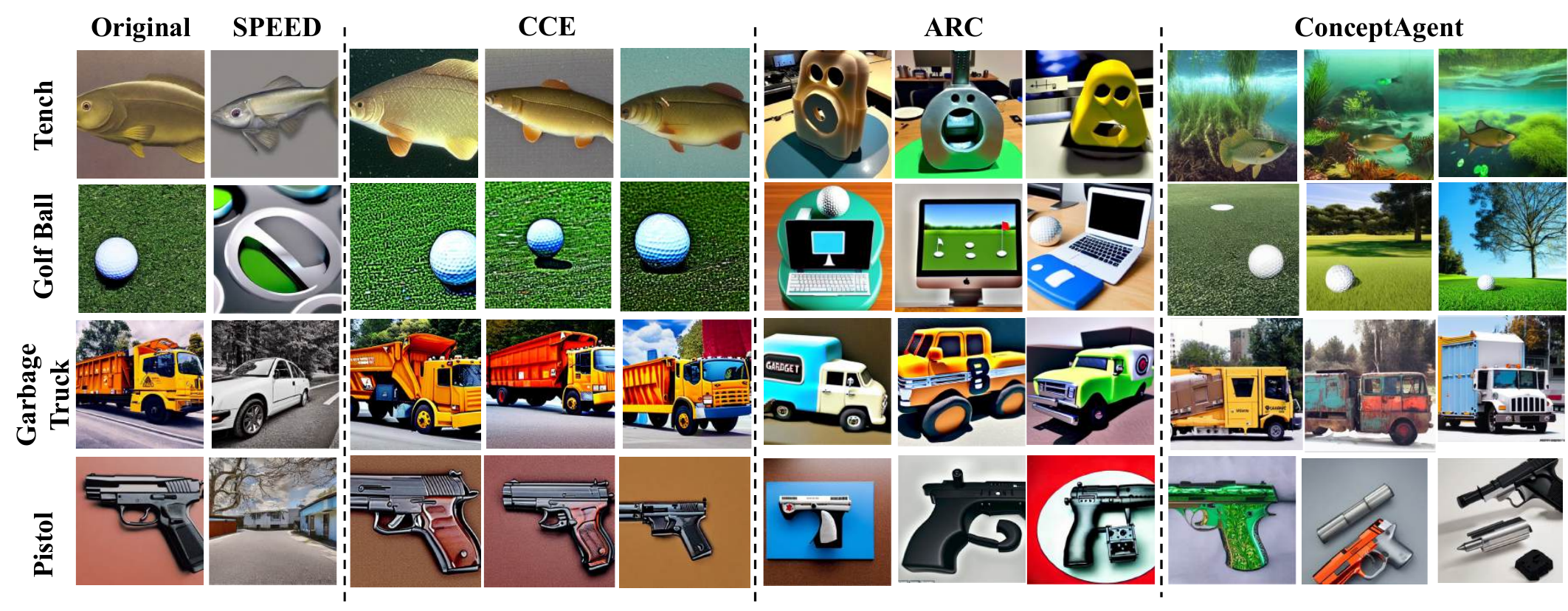}
        \captionof{figure}{Quantitative comparison of our proposed method against the training-based CCE and zero-shot training-free approaches ARC using ACC and CLIP score.}
        \label{erase-awake-mllm}
    \end{minipage}
    \hfill
    \begin{minipage}[t]{0.39\textwidth}
        \centering\includegraphics[width=\linewidth,height=5.0cm,keepaspectratio]{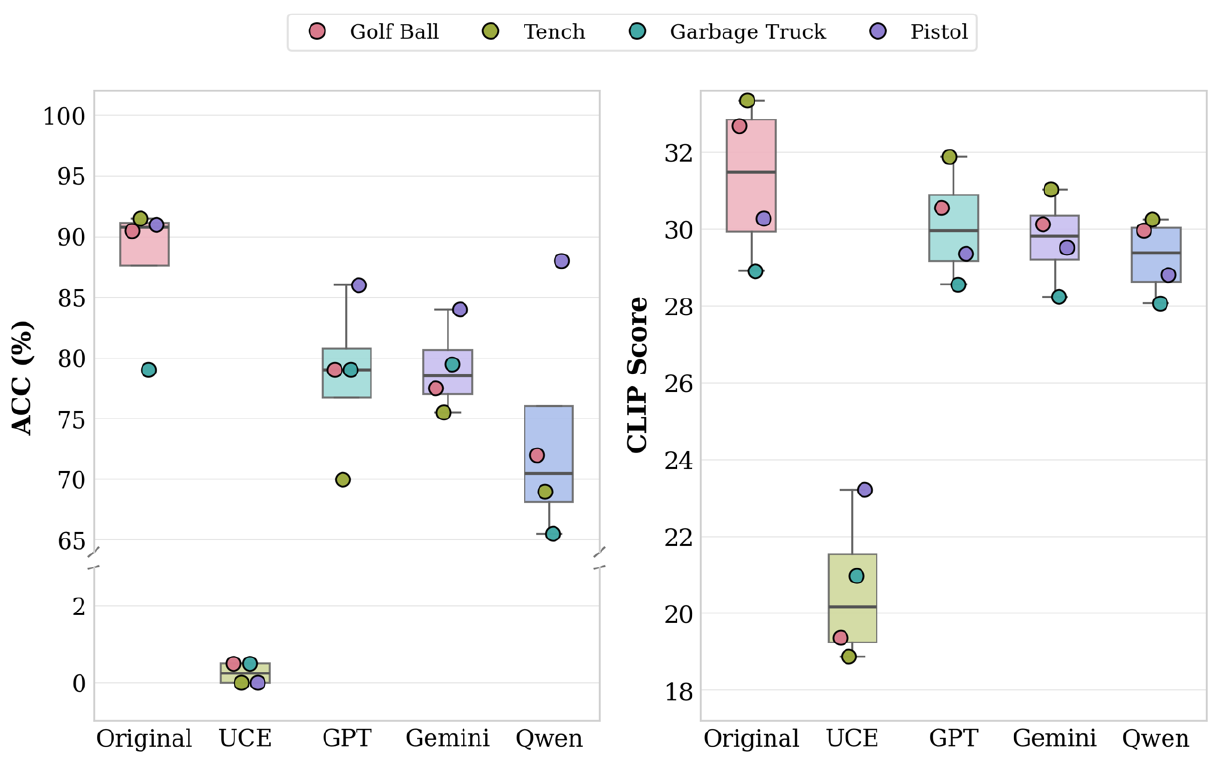}
        \captionof{figure}{Cross-model evaluation of ConceptAgent across diverse MLLMs and erased concepts.}
        \vspace{-15pt}
        \label{2.1}
    \end{minipage}
\vspace{-20pt}
\end{figure}

Quantitative results in Figure~\ref{erase-awake-mllm} further confirm the superiority of ConceptAgent. 
Compared to prior methods, our method achieves competitive or superior ACC and CLIP scores without requiring additional training or access to real concept images.
CCE inserts a small number of concept examples through targeted fine-tuning under white-box conditions. While effective, this approach resembles few-shot adaptation and often reduces diversity, as generated outputs tend to mimic the provided samples. Moreover, it requires access to model parameters and curated concept data, limiting its applicability.
ARC operates under a black-box setting but assumes alignment between semantic relationships and the generative trajectory. In practice, this assumption may not hold due to the highly entangled nature of diffusion representations, which leads to degraded generation quality and unstable awakening performance. 

In contrast, ConceptAgent operates in a black-box setting without additional training or real data. By leveraging trajectory-level intervention and surrogate-guided reasoning, it avoids strong assumptions about concept disentanglement and achieves more stable and visually coherent awakening results. 

\subsection{Generalization Analysis}
\begin{wrapfigure}{r}{0.5\linewidth}
\vspace{-10pt}
    \centering    \includegraphics[width=\linewidth]{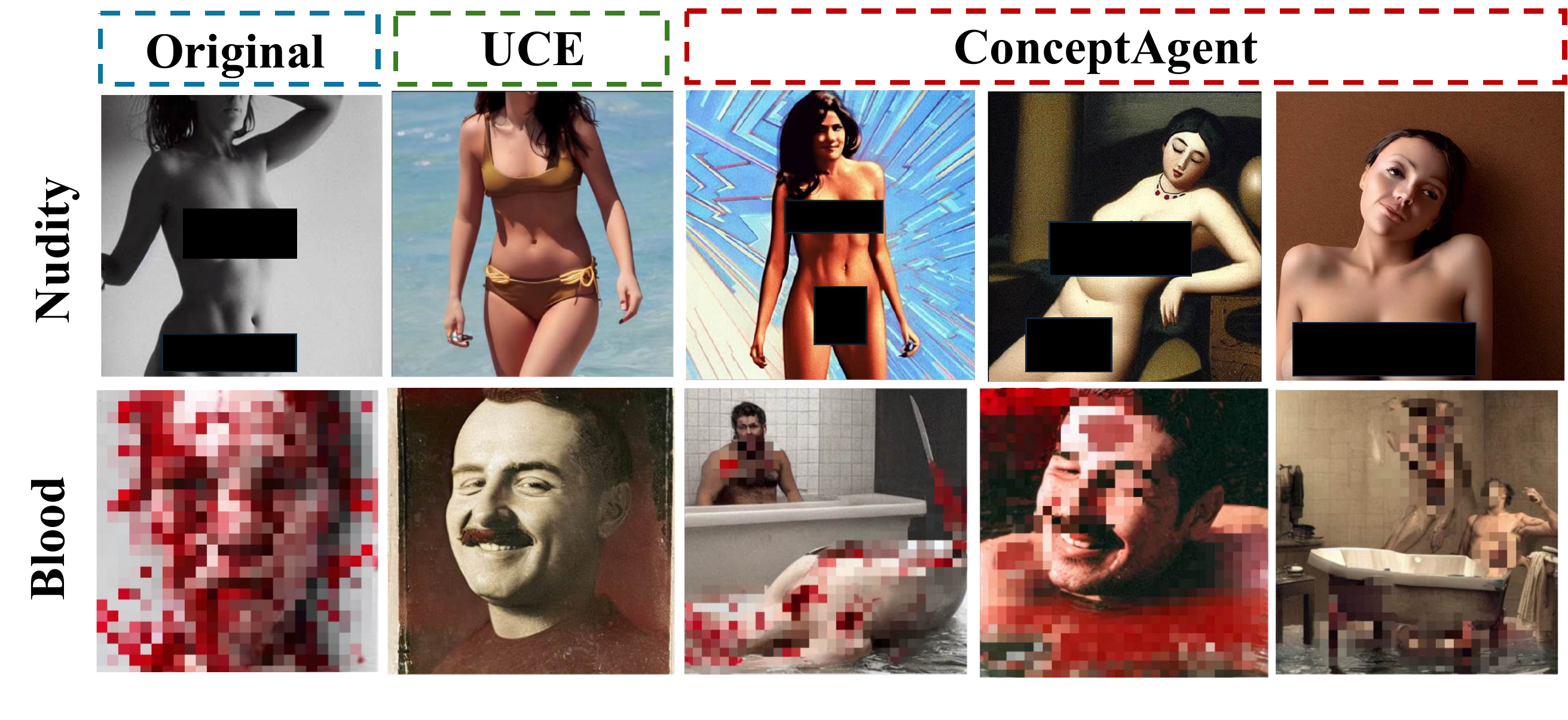}
    \vspace{-20pt}
    \caption{Qualitative Visualization of Harmful Concept Awakening using ConceptAgent.}
    \label{fig:harmful}
    \vspace{-10pt}
\end{wrapfigure}
To evaluate the generalizability of ConceptAgent across DMs, we further conduct experiments on SD~v2.1. 
As shown in Figure~\ref{2.1}, ConceptAgent consistently awakens erased concepts with impressive image quality across different MLLMs. 
This generalization ability stems from the fact that ConceptAgent operates at the trajectory level, rather than relying on model-specific components such as architecture, parameters, or internal representations. By intervening directly in the denoising process, our method remains agnostic to the underlying model design, enabling seamless transfer across different diffusion backbones without adaptation. 

We further extend the evaluation to safety-critical concepts, including nudity and blood, which are commonly targeted by concept erasure methods. As illustrated in Fig.~\ref{fig:harmful}, ConceptAgent successfully awakens these harmful concepts across. This result indicates that the vulnerability revealed by our framework is not limited to benign object categories, but also applies to safety-sensitive content.

\section{Conclusion}
\label{conclu}
In this work, we revisit concept erasure from a trajectory-based perspective and highlight a key vulnerability in existing methods. By treating generation as a static process and focusing primarily on disrupting text-concept associations, previous approaches fail to account for the dynamic evolution of semantic information along the denoising trajectory. Motivated by this insight, we propose \textit{ConceptAgent}, a training-free, zero-shot, black-box framework that achieves concept awakening by intervening in intermediate denoising states via surrogate-guided noise initialization. 
Our method operates a single inference pass, without requiring optimization or fine-tuning at any stage. Beyond effectiveness, ConceptAgent provides improved interpretability through its multi-agent design. Each agent performs a well-defined role using explicit semantic descriptions and structured visual priors, forming a transparent reasoning pipeline that contrasts with opaque embedding-space manipulation commonly used in prior approaches.
Extensive experiments demonstrate that ConceptAgent consistently awakens erased concepts across diverse DMs and erasure methods, while maintaining high visual quality and diversity. In future work, we aim to investigate more robust and principled defenses against trajectory-level vulnerabilities, and to develop generation frameworks that ensure safety under adversarial or manipulated inputs.

\bibliographystyle{plainnat}
\bibliography{ref}

@misc{zhang2024generatenotsafetydrivenunlearned,
      title={To Generate or Not? Safety-Driven Unlearned Diffusion Models Are Still Easy To Generate Unsafe Images ... For Now}, 
      author={Yimeng Zhang and Jinghan Jia and Xin Chen and Aochuan Chen and Yihua Zhang and Jiancheng Liu and Ke Ding and Sijia Liu},
      year={2024},
      eprint={2310.11868},
      archivePrefix={arXiv},
      primaryClass={cs.CV},
      url={https://arxiv.org/abs/2310.11868}, 
}

@article{pham2023circumventing,
  title={Circumventing concept erasure methods for text-to-image generative models},
  author={Pham, Minh and Marshall, Kelly O and Cohen, Niv and Mittal, Govind and Hegde, Chinmay},
  journal={arXiv preprint arXiv:2308.01508},
  year={2023}
}

@inproceedings{hsu2024ring,
  title={RING-A-BELL! HOW RELIABLE ARE CONCEPT REMOVAL METHODS FOR DIFFUSION MODELS?},
  author={Hsu, Chia Yi and Tsai, Yu Lin and Xie, Chulin and Lin, Chih Hsun and Chen, Jia You and Li, Bo and Chen, Pin Yu and Yu, Chia Mu and Huang, Chun Ying},
  booktitle={12th International Conference on Learning Representations, ICLR 2024},
  year={2024}
}

@inproceedings{lu2024mace,
  title={Mace: Mass concept erasure in diffusion models},
  author={Lu, Shilin and Wang, Zilan and Li, Leyang and Liu, Yanzhu and Kong, Adams Wai-Kin},
  booktitle={Proceedings of the IEEE/CVF Conference on Computer Vision and Pattern Recognition},
  pages={6430--6440},
  year={2024}
}

@article{xie2025erasing,
  title={Erasing Concepts, Steering Generations: A Comprehensive Survey of Concept Suppression},
  author={Xie, Yiwei and Liu, Ping and Zhang, Zheng},
  journal={arXiv preprint arXiv:2505.19398},
  year={2025}
}

@article{cao2025controllable,
  title={Controllable generation with text-to-image diffusion models: A survey},
  author={Cao, Pu and Zhou, Feng and Song, Qing and Yang, Lu},
  journal={IEEE Transactions on Pattern Analysis and Machine Intelligence},
  year={2025},
  publisher={IEEE}
}

@article{liu2025survey,
  title={A Survey on Cache Methods in Diffusion Models: Toward Efficient Multi-Modal Generation},
  author={Liu, Jiacheng and Wang, Xinyu and Lin, Yuqi and Wang, Zhikai and Wang, Peiru and Cai, Peiliang and Zhou, Qinming and Yan, Zhengan and Yan, Zexuan and Shi, Zhengyi and others},
  journal={arXiv preprint arXiv:2510.19755},
  year={2025}
}

@article{divya2024transforming,
  title={Transforming content creation: The influence of generative AI on a new frontier},
  author={Divya, V and Mirza, Agha Urfi},
  journal={Exploring the frontiers of artificial intelligence and machine learning technologies},
  volume={143},
  year={2024},
  publisher={San International Scientific Publication}
}

@article{grebe2025erased,
  title={Erased but Not Forgotten: How Backdoors Compromise Concept Erasure},
  author={Grebe, Jonas Henry and Braun, Tobias and Rohrbach, Marcus and Rohrbach, Anna},
  journal={arXiv preprint arXiv:2504.21072},
  year={2025}
}

@inproceedings{gandikota2024unified,
  title={Unified concept editing in diffusion models},
  author={Gandikota, Rohit and Orgad, Hadas and Belinkov, Yonatan and Materzy{\'n}ska, Joanna and Bau, David},
  booktitle={Proceedings of the IEEE/CVF Winter Conference on Applications of Computer Vision},
  pages={5111--5120},
  year={2024}
}

@inproceedings{gandikota2023erasing,
  title={Erasing concepts from diffusion models},
  author={Gandikota, Rohit and Materzynska, Joanna and Fiotto-Kaufman, Jaden and Bau, David},
  booktitle={Proceedings of the IEEE/CVF international conference on computer vision},
  pages={2426--2436},
  year={2023}
}

@inproceedings{orgad2023editing,
  title={Editing implicit assumptions in text-to-image diffusion models},
  author={Orgad, Hadas and Kawar, Bahjat and Belinkov, Yonatan},
  booktitle={Proceedings of the IEEE/CVF International Conference on Computer Vision},
  pages={7053--7061},
  year={2023}
}

@inproceedings{gong2024reliable,
  title={Reliable and efficient concept erasure of text-to-image diffusion models},
  author={Gong, Chao and Chen, Kai and Wei, Zhipeng and Chen, Jingjing and Jiang, Yu-Gang},
  booktitle={European Conference on Computer Vision},
  pages={73--88},
  year={2024},
  organization={Springer}
}

@inproceedings{lyu2024one,
  title={One-dimensional adapter to rule them all: Concepts diffusion models and erasing applications},
  author={Lyu, Mengyao and Yang, Yuhong and Hong, Haiwen and Chen, Hui and Jin, Xuan and He, Yuan and Xue, Hui and Han, Jungong and Ding, Guiguang},
  booktitle={Proceedings of the IEEE/CVF Conference on Computer Vision and Pattern Recognition},
  pages={7559--7568},
  year={2024}
}

@article{li2025speed,
  title={Speed: Scalable, precise, and efficient concept erasure for diffusion models},
  author={Li, Ouxiang and Wang, Yuan and Hu, Xinting and Jiang, Houcheng and Liang, Tao and Hao, Yanbin and Ma, Guojun and Feng, Fuli},
  journal={arXiv preprint arXiv:2503.07392},
  year={2025}
}

@inproceedings{rusanovsky2025memories,
  title={Memories of forgotten concepts},
  author={Rusanovsky, Matan and Malnick, Shimon and Jevnisek, Amir and Fried, Ohad and Avidan, Shai},
  booktitle={Proceedings of the Computer Vision and Pattern Recognition Conference},
  pages={2966--2975},
  year={2025}
}

@misc{howard2019imagenette,
  author       = {Jeremy Howard},
  title        = {Imagenette},
  year         = {2019},
  howpublished = {\url{https://github.com/fastai/imagenette/}}
}

@article{carion2025sam,
  title={Sam 3: Segment anything with concepts},
  author={Carion, Nicolas and Gustafson, Laura and Hu, Yuan-Ting and Debnath, Shoubhik and Hu, Ronghang and Suris, Didac and Ryali, Chaitanya and Alwala, Kalyan Vasudev and Khedr, Haitham and Huang, Andrew and others},
  journal={arXiv preprint arXiv:2511.16719},
  year={2025}
}

@article{dhariwal2021diffusion,
  title={Diffusion models beat gans on image synthesis},
  author={Dhariwal, Prafulla and Nichol, Alexander},
  journal={Advances in neural information processing systems},
  volume={34},
  pages={8780--8794},
  year={2021}
}

@article{lu2025concepts,
  title={When Are Concepts Erased From Diffusion Models?},
  author={Lu, Kevin and Kriplani, Nicky and Gandikota, Rohit and Pham, Minh and Bau, David and Hegde, Chinmay and Cohen, Niv},
  journal={arXiv preprint arXiv:2505.17013},
  year={2025}
}

@inproceedings{qu2023unsafe,
  title={Unsafe diffusion: On the generation of unsafe images and hateful memes from text-to-image models},
  author={Qu, Yiting and Shen, Xinyue and He, Xinlei and Backes, Michael and Zannettou, Savvas and Zhang, Yang},
  booktitle={Proceedings of the 2023 ACM SIGSAC conference on computer and communications security},
  pages={3403--3417},
  year={2023}
}

@article{kim2025comprehensive,
  title={A comprehensive survey on concept erasure in text-to-image diffusion models},
  author={Kim, Changhoon and Qi, Yanjun},
  journal={arXiv preprint arXiv:2502.14896},
  year={2025}
}

@article{liu2025erased,
  title={Erased or Dormant? Rethinking Concept Erasure Through Reversibility},
  author={Liu, Ping and Zhang, Chi},
  journal={arXiv preprint arXiv:2505.16174},
  year={2025}
}

@article{sun2026lure,
  title={LURE: Latent Space Unblocking for Multi-Concept Reawakening in Diffusion Models},
  author={Sun, Mengyu and Yang, Ziyuan and Teoh, Andrew Beng Jin and Liu, Junxu and Hu, Haibo and Zhang, Yi},
  journal={arXiv preprint arXiv:2601.14330},
  year={2026}
}

@article{chen2025ghostprompt,
  title={GhostPrompt: Jailbreaking Text-to-image Generative Models based on Dynamic Optimization},
  author={Chen, Zixuan and Lin, Hao and Xu, Ke and Jiang, Xinghao and Sun, Tanfeng},
  journal={arXiv preprint arXiv:2505.18979},
  year={2025}
}

@article{weng2025m,
  title={M-ErasureBench: A Comprehensive Multimodal Evaluation Benchmark for Concept Erasure in Diffusion Models},
  author={Weng, Ju-Hsuan and Liao, Jia-Wei and Chou, Cheng-Fu and Chen, Jun-Cheng},
  journal={arXiv preprint arXiv:2512.22877},
  year={2025}
}

@inproceedings{lee2025localized,
  title={Localized concept erasure for text-to-image diffusion models using training-free gated low-rank adaptation},
  author={Lee, Byung Hyun and Lim, Sungjin and Chun, Se Young},
  booktitle={Proceedings of the Computer Vision and Pattern Recognition Conference},
  pages={18596--18606},
  year={2025}
}

@inproceedings{petsiuk2024concept,
  title={Concept arithmetics for circumventing concept inhibition in diffusion models},
  author={Petsiuk, Vitali and Saenko, Kate},
  booktitle={European Conference on Computer Vision},
  pages={309--325},
  year={2024},
  organization={Springer}
}

@article{gorgun2025temporal,
  title={Temporal Concept Dynamics in Diffusion Models via Prompt-Conditioned Interventions},
  author={Gorgun, Ada and Sammani, Fawaz and Deligiannis, Nikos and Schiele, Bernt and Fischer, Jonas},
  journal={arXiv preprint arXiv:2512.08486},
  year={2025}
}

@article{yang2025qwen3,
  title={Qwen3 technical report},
  author={Yang, An and Li, Anfeng and Yang, Baosong and Zhang, Beichen and Hui, Binyuan and Zheng, Bo and Yu, Bowen and Gao, Chang and Huang, Chengen and Lv, Chenxu and others},
  journal={arXiv preprint arXiv:2505.09388},
  year={2025}
}

@article{singh2025openai,
  title={Openai gpt-5 system card},
  author={Singh, Aaditya and Fry, Adam and Perelman, Adam and Tart, Adam and Ganesh, Adi and El-Kishky, Ahmed and McLaughlin, Aidan and Low, Aiden and Ostrow, AJ and Ananthram, Akhila and others},
  journal={arXiv preprint arXiv:2601.03267},
  year={2025}
}

@article{comanici2025gemini,
  title={Gemini 2.5: Pushing the frontier with advanced reasoning, multimodality, long context, and next generation agentic capabilities},
  author={Comanici, Gheorghe and Bieber, Eric and Schaekermann, Mike and Pasupat, Ice and Sachdeva, Noveen and Dhillon, Inderjit and Blistein, Marcel and Ram, Ori and Zhang, Dan and Rosen, Evan and others},
  journal={arXiv preprint arXiv:2507.06261},
  year={2025}
}

@inproceedings{he2016deep,
  title={Deep residual learning for image recognition},
  author={He, Kaiming and Zhang, Xiangyu and Ren, Shaoqing and Sun, Jian},
  booktitle={Proceedings of the IEEE conference on computer vision and pattern recognition},
  pages={770--778},
  year={2016}
}

@inproceedings{hessel2021clipscore,
  title={Clipscore: A reference-free evaluation metric for image captioning},
  author={Hessel, Jack and Holtzman, Ari and Forbes, Maxwell and Le Bras, Ronan and Choi, Yejin},
  booktitle={Proceedings of the 2021 conference on empirical methods in natural language processing},
  pages={7514--7528},
  year={2021}
}

@article{kwon2022diffusion,
  title={Diffusion models already have a semantic latent space},
  author={Kwon, Mingi and Jeong, Jaeseok and Uh, Youngjung},
  journal={arXiv preprint arXiv:2210.10960},
  year={2022}
}



\end{document}